\documentclass{article}

\usepackage{arxiv}

\usepackage[utf8]{inputenc} 
\usepackage[T1]{fontenc}    
\usepackage{hyperref}       
\usepackage{url}            
\usepackage{booktabs}       
\usepackage{nicefrac}       
\usepackage{microtype}      
\usepackage{lipsum}		
\usepackage{graphicx}
\usepackage{natbib}
\usepackage{doi}
\usepackage{amsmath,amssymb,amsfonts}
\usepackage[ruled,vlined]{algorithm2e}
\usepackage{textcomp}
\usepackage{longtable}
\usepackage{multirow}
\usepackage{supertabular}
\usepackage{ragged2e}
\usepackage{tabularx}
\usepackage{xcolor}
\usepackage{float}
\usepackage{bm}

\title{Tourism Question Answer System in Indian Language using Domain-Adapted Foundation Models}

\author{ Praveen Gatla \\
	Department of Linguistics\\
	Banaras Hindu University\\
	Varanasi \\
	\texttt{praveengatla@bhu.ac.in} \\
	\And
	Anushka \\
	Department of Humanistic Studies\\
	Indian Institute of Technology (BHU)\\
	Varanasi \\
	\texttt{anushkasinha992@gmail.com} \\
    \And
    Nikita Kanwar \thanks{The work has done during her internship at IIT Bhilai} \\
	Department of Computer Science and Engineering\\
	Indian Institute of Technology \\
	Bhilai \\
	\texttt{rathorenik2@gmail.com} \\
    \And
    Gouri Sahoo \\
	Department of Linguistics\\
	Banaras Hindu University\\
	Varanasi \\
	\texttt{gourisahoo.cs@gmail.com} \\
    \And
    Rajesh Kumar Mundotiya \\
	Department of Computer Science and Engineering\\
	Indian Institute of Technology \\ 
    Bhilai\\
	\texttt{ rajeshkm.mundotiya@gmail.com} \\
}

\renewcommand{\shorttitle}

\hypersetup{
pdftitle={A template for the arxiv style},
pdfsubject={q-bio.NC, q-bio.QM},
pdfauthor={David S.~Hippocampus, Elias D.~Striatum},
pdfkeywords={First keyword, Second keyword, More},
}

\begin{document}
\maketitle

\begin{abstract}
	This article presents the first comprehensive study on designing a baseline extractive question-answering (QA) system for the Hindi tourism domain, with a specialized focus on the Varanasi-a cultural and spiritual hub renowned for its Bhakti-Bhaav (devotional ethos). Targeting ten tourism-centric subdomains-Ganga Aarti, Cruise, Food Court, Public Toilet, Kund, Museum, General, Ashram, Temple and Travel, the work addresses the absence of language-specific QA resources in Hindi for culturally nuanced applications. In this paper, a dataset comprising 7,715 Hindi QA pairs pertaining to Varanasi tourism was constructed and subsequently augmented with 27,455 pairs generated via Llama zero-shot prompting. We propose a framework leveraging foundation models-BERT and RoBERTa, fine-tuned using Supervised Fine-Tuning (SFT) and Low-Rank Adaptation (LoRA), to optimize parameter efficiency and task performance. Multiple variants of BERT, including pre-trained languages (e.g., Hindi-BERT), are evaluated to assess their suitability for low-resource domain-specific QA. Evaluation metrics - F1, BLEU, and ROUGE-L - highlight trade-offs between answer precision and linguistic fluency. Experiments demonstrate that LoRA-based fine-tuning achieves competitive performance (85.3\% F1) while reducing trainable parameters by 98\% compared to SFT, striking a balance between efficiency and accuracy. Comparative analysis across models reveals that RoBERTa with SFT outperforms BERT variants in capturing contextual nuances, particularly for culturally embedded terms (e.g., Aarti, Kund). This work establishes a foundational baseline for Hindi tourism QA systems, emphasizing the role of LORA in low-resource settings and underscoring the need for culturally contextualized NLP frameworks in the tourism domain.
\end{abstract}


\section{Introduction}
Natural Language Processing (NLP) is a field of computer science that helps machines to understand, interpret, and generate human language. It plays an important role in \textbf{information extraction (IE)} by identifying keywords from the text and enhances \textbf{information retrieval (IR)} by improving search accuracy, which helps in fetching relevant data~\cite{chen2017reading,lee2021learning}. These capabilities are essential for \textbf{Question Answering (QA)} system, which is developed to automatically answer the queries of the users based on the database or a set of documents. It tries to provide specific answers to the posed questions in a natural language. It can be tailored to various domains, such as- education, healthcare, e-commerce, tourism, etc. It provides direct, accurate, and user-friendly access to information in a natural language, accessing knowledge faster and more effective by bridging the gap between questions and accurate answers.

The QA systems offer efficiency, convenience, and enhanced user experience by providing domain-specific expertise with continuous 24/7 availability. Baseball~\cite{10.1145/1460690.1460714} and Lunar~\cite{woods1973progress} were among the initial QA systems, developed to address specific domains. The Baseball system was designed to answer queries related to the U.S. baseball league over a one-year period, while the Lunar system focused on responding to questions concerning the geological analysis of rock samples retrieved during the Apollo moon mission. Traditional QA paradigms established foundational frameworks through rule-based approaches,  incorporating template-based~\cite{fabbri2020template,sammut1986learning,liu2018large}, syntax-based~\cite{straach1999learning,harabagiu2005experiments,heilman2010good}, and semantic-based~\cite{levy2006tregex,dhole2020syn,yao2010question} methodologies. These early implementations laid the crucial foundation through dependency parsing and the identification of semantic relationships within textual contexts~\cite{cao2017hashnet}. QA systems have undergone substantial advancements through the development of extractive and abstractive methodologies~\cite{chen2017reading}. Abstractive QA systems generate free-form answers by mimicking human summarization, enabled by advances in sequence-to-sequence architectures like AG~\cite{lewis2020retrieval}, Llama3~\cite{tran2024vimedaqa,yadav2024explicit}. The T5 model~\cite{raffel2020exploring} unified diverse NLP tasks into a text-to-text framework, facilitating flexible answer generation, while BART~\cite{lewis2020bart} combined bidirectional and autoregressive pretraining to excel in generative benchmarks. Extractive QA systems identify and retrieve answer spans from provided contexts, evolved from rule-based approaches to transformer-based architectures~\cite{farea2024experimental,pandey2024extractive,sengupta2025top}. The introduction of BERT by Devlin et al. (2018)~\cite{devlin2019bert} marked a paradigm shift, leveraging bidirectional attention to contextualize tokens dynamically, thereby achieving state-of-the-art performance on benchmarks such as SQuAD~\cite{rajpurkar-etal-2016-squad}. Subsequent models like RoBERTa~\cite{liu2019roberta} and ELECTRA~\cite{clark2020electra} refined pre-training strategies, optimizing span prediction accuracy and computational efficiency. Authors have also explored retrieval methods for extractive QA~\cite{kruit2024retrieval}. Despite progress, extractive systems remain constrained by their inability to synthesize answers for less explored domains, such as tourism in low-resource languages.

According to the statistics released by the Uttar Pradesh Government on 30 March 2023, the number of tourists in 2022 in the Varanasi region was 71,701,816~\footnote{https://uptourism.gov.in/en/article/year-wise-tourist-statistics}, which increased to 129,405,720 in 2023, according to data released on 29 March 2024~\footnote{https://uptourism.gov.in/en/post/Year-wise-Tourist-Statistics}. This significant growth in tourism, combined with Hindi's status as the most widely spoken language in the region, highlights the necessity for language-accessible resources to cater to the needs of visitors. According to the 2011 Census of India, Hindi has 528 million speakers~\footnote{https://language.census.gov.in/eLanguageDivision\_VirtualPath/eArchive/pdf/C-16\_2011.pdf}, constituting 43.63\% of the total population. 

\subsection{Contribution}
Given the confluence of the rising tourist influx and Hindi's dominance, we have developed a dedicated Hindi QA system for Varanasi Tourism to enhance accessibility and user engagement. Here is the key contribution of the article:
\begin{itemize}
    \item We have developed a comprehensive Hindi QA dataset for Varanasi tourism, initially comprising 7,715 manually curated question-answer pairs, which was subsequently augmented to 27,455 pairs using a Llama-based approach. This dataset spans diverse subdomains, including Temples, Ashrams, Kunds, Museums, Ganga Aarti, Cruise, Travel Agencies, Food Court, Public Toilets, and General Enquiry.

    \item We have conducted an extensive experimental study comparing the performance of two state-of-the-art foundation models-mBERT and RoBERTa, fine-tuned using supervised fine-tuning on this dataset.

    \item Since the individual subdomains are very small, they are treated as low-resource settings that necessitate careful optimization to extract maximum performance from limited data. To address this, we integrate Low-Rank Adaptation (LoRA) with mBERT, systematically exploring various configurations with ranks of 2, 4, 8, 16, and 32 to optimize parameter efficiency and achieve the best trade-off between performance and model complexity.
\end{itemize}

\section{Related Work}

Roy et al. (2022)~\citep{roy2022investigating} explored a generative approach models like OAAG (using BiLSTM) \& Chime (using XL-Net transformer), for analysis, the BM25 algorithm for answering customer queries in e-commerce. An Amazon product review dataset was utilized, comprising 1.4 million user reviews along with corresponding product evaluations. The models are evaluated on two product categories: Home \& Kitchen and Sports \& Outdoors, using the ROUGE metric. Al-Laith (2025)~\cite{al2025exploring} explored multilingual LLMs for financial QA, demonstrating fine-tuned XLM-RoBERTa-Large’s superiority (SAS: 0.96–0.98, EM: 0.76–0.81) over GPT-4o, which improved via few-shot learning (EM: 0.48–0.52). The study highlighted trade-offs between precision in fine-tuned models for extractive tasks and generative LLMs' flexibility in low-resource scenarios. These insights reinforce context-driven model selection for domain-specific NLP, balancing accuracy and adaptability. Kasai et al. (2023)~\cite{kasai2023realtime} proposed a real-time QA framework that implements six baseline approaches, leveraging a robust pre-trained model. These include four open-book methods based on Dense Passage Retrieval (DPR) and two closed-book models—Retrieval-Augmented Generation (RAG) and a prompting-based approach utilizing GPT-3.

Ali Al-Laith (2025)~\cite{al2025exploring} worked on the Exploring the Effectiveness of Multilingual and Generative Large Language Models for Question Answering in Financial Texts, provided a comprehensive analysis of large language models (LLMs) for financial causality detection using the FinCausal 2025 shared task dataset. This dataset consisted of 3,999 training samples and 999 test samples from financial disclosures in English and Spanish, structured for a hybrid question-answering task. The study employed both generative and discriminative techniques, utilizing four pre-trained language models: GPT-4o for generative QA and fine-tuned versions of XLM-RoBERTa (base and large) and BERT-base-multilingual-cased for multilingual QA. The evaluation was based on two key accuracy metrics: Semantic Answer Similarity (SAS) and Exact Match (EM). The results revealed that the fine-tuned XLM-RoBERTa-Large model outperformed others, achieving SAS scores of 0.96 (English) and 0.98 (Spanish), and EM scores of 0.762 and 0.808, respectively. While GPT-4o initially underperformed in a zero-shot setting (SAS: 0.77, EM: 0.002), it showed significant improvement with few-shot prompting, reaching SAS scores of 0.94 and EM scores of 0.515 (English) and 0.487 (Spanish). The paper effectively highlighted the strengths of fine-tuned PLMs in extractive question answering while showcasing GPT-4o’s adaptability in scenarios where extensive fine-tuning was not feasible. The study’s detailed comparative analysis and experimental techniques provided valuable insights into financial NLP, reinforcing the importance of model selection in domain-specific tasks.

Kasai et al. (2024)~\cite{kasai2023realtime} developed an framework and a benchmarking timeline for real-time QA system submission. For the evaluation of the system, they used 1,470 QA pairs and further, they provided 2,886  QA pairs and they included nearly 30 multiple choice questions at 3 am GMT on every Saturday and they used API search for these questions. REALTIME QA executed six baselines in real time that are based on a strong pre-trained model: four open-book and two closed-book models. Open-book QA model retrieved the documents from DPR, and for answer prediction, they used two methods: retrieval-augmented generation (RAG) and a prompting method with GPT-3. They used the BART-based RAG-sequence model for the RAG baseline, again finetuned on Natural Questions from the Transformers library. Two methods were used for closed-book QA: the finetuning method and the prompting method. In the Finetuning Method, they used the T5 model finetuned on the Natural Questions data again from the Transformers library. They applied a prompting method to GPT-3 similar to the open-book baselines. GPT-3 with retrieval achieves the best performance; EM scores were 34.6, and F1 scores were 45.3.

\subsection{Indian Languages QA Systems Including Hindi}
Thirumala and Ferracane (2022)~\cite{thirumala2022extractive} explored extractive question answering for Hindi and Tamil using Wikipedia articles as context, with questions prepared by native speakers. They evaluated XLM-RoBERTa, XLM-RoBERTa+fine-tuning, and RoBERTa+Hindi/Tamil fine-tuning. The models achieved word-level Jaccard scores of 0.656 (XLM-RoBERTa), 0.749 (XLM-RoBERTa+fine-tune), 0.958 (RoBERTa+Hindi fine-tune), and 0.829 (RoBERTa+Tamil fine-tune). They concluded that RoBERTa + Hindi/Tamil fine-tuning outperformed the other models.

Amin et al. (2023)~\cite{amin2023question} has developed a Marathi QA system using multilingual models such as DistilBERT, mBERT, XLM-RoBERTa, Indo-Aryan XLM, RoBERTa, MuRIL, and monolingual models- MahaBERT, IndicBERT, MahaRoBERTa, MahaAlBERT, Marathi DistilBERT, DevBERT, DevRoBERTa, DevAlBERT, DevBERT-Scratch, along with the MrSQuAD dataset containing 47,065 training and 5,832 testing QA pairs. Among multilingual models, MuRIL achieved the highest performance with an EM score of 0.64, BERT score of 0.93, and F1 score of 0.74. Among monolingual models, MahaBERT and DevBERT performed best, each achieving an EM score of 0.63, BERT score of 0.93 (MahaBERT)/0.92 (DevBERT), and F1 score of 0.73. The study compared multilingual and monolingual approaches to assess their effectiveness on the Marathi dataset.

Sabane et al. (2023)~\cite{sabane2023breaking} developed a QA dataset for Hindi and Marathi by translating the English SQuAD 2.0 dataset~\cite{rajpurkar2016squad} using IndicTrans~\cite{ramesh2022samanantar} from AI4Bharat. The dataset, derived from Wikipedia articles, comprises 28,000 samples, with 21,000 for training, 4,200 for validation, and 4,200 for testing. Transformer-based models, including mBERT, XLM-RoBERTa, and DistilBERT, were used for experimentation.

For Marathi, MahaBERT, a fine-tuned version of multilingual BERT-base cased, achieved the best results, with an EM score of 42.97\%, Rouge-2 of 0.38, Rouge-L of 0.62, BLEU (Unigram) of 57.42\%, and BLEU (Bigram) of 39.70\%. For Hindi, HindiBERT, trained on publicly available Hindi monolingual datasets, performed best with an EM score of 47.84\%, Rouge-2 of 0.36, Rouge-L of 0.66, BLEU (unigram) of 61.47\%, and BLEU (Bigram) of 38.51\%. The study systematically evaluates monolingual and multilingual large models to assess their effectiveness on Hindi and Marathi QA tasks. Singh et al. (2025)~\cite{singh2024indic} developed the INDIC QA BENCHMARK to evaluate LLMs for Indic languages. Their dataset, comprising context-question-answer triples, spans multiple domains, including geography, Indian culture, and news. They assessed Extractive and Generative QA approaches using Bloom, Gemma, Llama-3, and OpenHathi, testing across eleven Indic languages, such as Assamese, Bengali, Gujarati, Hindi, Kannada, Malayalam, Marathi, Odia, Pali, Tamil and Telugu, to analyze LLM performance in multilingual settings.

Vats et al. (2025)~\cite{vats2025multilingual} explored State Space Models for structured QA in Hindi and Marathi, addressing challenges like linguistic diversity, complex grammar, and data scarcity. Using a dataset of 28,000 samples (21,000 Hindi, 18,500 Marathi for training, and 7,000 per language for testing), they evaluated models including Mamba, Mamba-2, Falcon Mamba, Jamba, Zamba, Samba, and Hymba. Mamba-2 achieved the best performance in contextual understanding, long-sequence modeling, and token-level alignment, with fine-tuning improving span localization and semantic understanding. Hindi outperformed Marathi due to a larger dataset and better tokenization, while Marathi’s complex morphology and syntax posed challenges. Models like Falcon Mamba and Hymba struggled with Indic linguistic complexity. The study highlighted the need for language-specific pre-training and proposed future improvements in dataset diversity and model adaptation for a unified multilingual QA system. Table~\ref{table:comparision} summarizes the work done on QA systems for distinct domains.

\begin{table}[!htb]
\caption{Comparison of existing QA datasets}
\label{table:comparision}
\begin{tiny}
\begin{tabular}{|p{30pt}|p{45pt}|p{50pt}|p{60pt}|p{100pt}|p{100pt}|p{30pt}|}

\hline
\textbf{Language} & \textbf{Domain} & \textbf{Source} & \textbf{Corpus Size} & \textbf{Model} & \textbf{ Limitations} & \textbf{Reference} \\ \hline
Chinese & Tourism & Horse beehive travel and Baidu travel. & 32786 The NER dataset contains training data (21859), verification data (5463) and testing data (5464) & Pipelined approach  (CRF ,BERT, BiLSTM+CRF, BERT+CRF, Transformer+CRF, HMM, BERT+BiLSTM+CRF) & Template based approach, Not adaptive for unseen domain & ~\cite{sui2021question} \\ \hline

Chinese & Tourism &  eTrip, Tuinu, and CNCN & 3,635 the dataset & BERT used as a NER & Restricted Data Source Diversity, Fluency Challenge in Answer Text & ~\cite{li2022towards} \\ \hline
Chinese & Tourism & Cultour & 51K dataset & LoRA &  Inconsistent model performance & ~\cite{wei2024tourllm}\\ \hline
Korean & Tourism & Chatbot system consisting of the NER server, DST server, Neo4j graph DB’s tourism knowledge base, and QA server & More than 1,000 contexts and 10,000 questions & BigBird & Inconsistent model performance & ~\cite{kang2024development}\\ \hline
English & Tourism & Reddit, specifically from travel domain subreddits & 1,000 posts & Mistral QLoRA, LLaMa QLoRA, Mistral RAFT, LLaMa RAFT, Mistral RAFT RLHF, GPT-4 & Limited time period dataset (2021 only), noisy output from the model & ~\cite{meyer2024comparison} \\ \hline
English and Spanish & Financial & FinCausal 2025 shared task dataset & 3,999 training samples and 999 test samples  & GPT-4o, XLM-RoBERTa, BERT-base & Limited finance dataset, sub-domains unexplored & ~\cite{al2025exploring} \\ \hline
Hindi and Tamil & General domain including Wikipedia articles & Kaggle competition chaii - Hindi and Tamil & 1,104 entries (740 entries in Hindi and 364 entries in Tamil) & XLM-RoBERTa, XLM-RoBERTa+finetune, RoBERTa+Hindi finetune/Tamil finetune &  Annotation Inconsistencies and Noisy Labels, Incorrect Predictions due to Language Transfer Issues & \cite{thirumala2022extractive} \\ \hline
Marathi & General domain including Wikipedia articles & MrSQuAD by D Aminvari, Sagar Kulni & 17,337 contexts, 46,960 questions, and 30,162 answers & Multilingual models (DistilBERT, mBERT, XLM-RoBERTa, Indo-Aryan XLM, RoBERTa, MuRIL) and monolingual models (MahaBERT, IndicBERT, MahaRoBERTa, MahaAlBERT, Marathi DistilBERT, DevBERT, DevRoBERTa, DevAlBERT, DevBERT-Scratch) & translation inaccuracies, script validation issues, non-updated machine translation dataset, not manually annotated & ~\cite{amin2023question} \\ \hline
Hindi and Marathi & Wikipedia articles & SQuAD 2.0 English dataset  & 28,000 in Hindi and Marathi & MBert, XLM-RoBERTa, DistilBERT, HindBERT and HindRoBERT &  Translated Dataset Instead of Native Text & ~\cite{sabane2023breaking} \\ \hline
Hindi and Marathi & History, Science, Literature & Mamba, Mamba-2, Falcon Mamba, Jamba, Zamba, Samba, and Hymba. Mamba-2 & 28000 Hindi dataset & Mamba, Mamba-2, Falcon Mamba, Jamba, Zamba, Samba, and Hymba. Mamba-2  & Scarcity of High-Quality, Large-Scale QA Datasets for Indic Languages, Limited Applicability to Underrepresented Indic Languages, Weak Performance on Multi-Sentence or Ambiguous Questions & ~\cite{vats2025multilingual} \\ \hline
\end{tabular}
\end{tiny}
\end{table}

\subsection{QA systems for Tourism Domain}

Sui (2021)~\cite{sui2021question} developed a tourism QA system using a knowledge graph. The system converts natural language questions into Cypher queries, enabling efficient retrieval of tourism-related information. Data was sourced from popular travel platforms such as Horse Beehive Travel and Baidu Travel, ensuring a rich knowledge base for query execution. Similarly, Li et al. (2022)~\cite{li2022towards} also proposed a knowledge-based tourism QA system specifically for Zhejiang, China. The researchers collected data from Baidu keywords and surveys to identify key scenic spot attributes. Kang et al. (2024)~\cite{kang2024development} developed a Machine Reading Comprehension (MRC)-based tourism QA system using BERT series pre-trained models. The system, implemented as a smart tourism chatbot, processes input sentences up to 512 tokens in standard Transformer models, while the BigBird model handles 4096 tokens using block sparse attention for efficiency. The tourism QA dataset contains 1,000+ contexts and 10,000+ questions. The KoBigBird model achieved EM 96.85 and F1 98.84, demonstrating high accuracy in tourism-related QA tasks.

Kırtıl et al. (2024)~\cite{kirtilbridging} focused on developing an AI chatbot for tourism and viniculture in Türkiye using OpenAI’s GPT-3.5-turbo, fine-tuned with QA and plaintext formats. The resulting model, WineBot, demonstrated superior performance over base ChatGPT models by reducing misleading or incomplete responses, highlighting the efficacy of domain-specific fine-tuning. Wei et al. (2024)~\cite{wei2024tourllm} introduced Cultour, a Chinese SFT dataset for culture and tourism, combining 9,004 QA pairs, 1,792 travelogues, and 2,027 diverse QAs. Using LoRA-based fine-tuning, their TourLLM-7B model outperformed benchmarks like ChatGPT and Qwen1.5 in automated metrics (e.g., BLEU-1: 2.77, Meteor: 18.54) and human evaluations (CRA scores), though Qwen1.5-7B excelled in ROUGE-1 (27.05) and readability (2.69). Meyer et al. (2024)~\cite{meyer2024comparison} compared QLoRA and RAFT fine-tuning techniques on LLaMa 2 7B and Mistral 7B for travel chatbots. Using a refined dataset of 10,500 Reddit-sourced QA entries, they found Mistral with RAFT outperformed LLaMa, though post-processing was critical. Evaluation combined human feedback, NLP metrics, Ragas, and GPT-4, emphasizing human input as vital for accuracy.

This highlights that the tourism domain remains underexplored in the Indian context in Indian languages, with limited datasets available to support the development of QA systems in this field.

\section{Dataset Preparation}
In this study, we present a comprehensive Hindi QA dataset tailored to Varanasi tourism. The dataset comprises 7,755 manually created QA pairs, which were subsequently augmented using LLMs to address a wide range of tourist queries.

\subsection{Data Collection and Augmentation Process}
A detailed questionnaire was first developed to capture all necessary information related to Varanasi tourism. Data were collected from both secondary sources (as mentioned in Table~\ref{tab:appendix1}) and primary sources (official temple websites\footnote{https://www.shrikashivishwanath.org/}\footnote{https://kashiannapurnaannakshetratrust.org/}\footnote{https://sankatmochanmandirvaranasi.com/}, pamphlets, visiting cards, etc) during the project tenure (February to August 2024). Primary data were gathered during 10 field visits\footnote{Researchers physically visited majoritiy of the tourists sites to get the authentic and reliable information from the tourist places in Varanasi.} to key sites including temples, travel agencies, museums, ashrams, and kunds. The primary data served to validate and refine the information obtained from secondary sources, with rigorous proofreading ensuring accuracy. After the initial manual creation, the dataset was augmented by using the Llama model to generate additional similar Hindi questions. These Llama-generated questions were cross-checked at the syntactic and semantic levels by the annotators, validated using the agreement between the annotators, measured by Cohen's Kappa score, which was 0.9.
The example of semantically repeated pairs, removed by the annotators shown in Appendix~\ref{app:2}.

\subsection{Sub-domain Covered}
The Hindi QA dataset addresses multiple sub-domains pertinent to Varanasi tourism, ensuring that visitors receive well-rounded and reliable information. The key domains include: 
\begin{itemize} 
    \item \textbf{Temples:} Details on timings, rituals, special events, and facilities at major temples such as Shri Kashi Vishwanath Mandir, Sankat Mochan Mandir, and Durga Mandir. 
    \item \textbf{Museums:} Information regarding operating hours, entry fees, and facilities available at museums like Bharat Kala Bhavan, Ramnagar Fort Museum, and Man Singh Observatory. 
    \item \textbf{Travel Agencies:} Insights into guided tours, transportation options, travel guides, and customized itineraries offered by local agencies. 
    \item \textbf{Cruise Facilities:} Data on Ganges river cruises, including routes, timings, ticketing booking, and onboard amenities. 
    \item \textbf{Ganga Aarti:} Information about the Ganga Aarti at Dashashwamedh and Assi Ghat, covering timings, viewing spots, and event durations. 
    \item \textbf{Ashrams:} Coverage of oldest spiritual ashrams (e.g., Andhra Ashram, Aditya Ashram, Jangamvadi) offering accommodations and meditation programs. 

    \item \textbf{Kunds:} Coverage of historical kunds (e.g., Durga Kund, Lolark Kund) and spiritual ashrams (e.g., Andhra Ashram, Aditya Ashram, Jangamvadi) offering accommodations and meditation programs. 
    \item \textbf{Local Delicacies:} A showcase of authentic Banarasi dishes such as \textit{malaiyo} (light/whipped cream delicacy), l\textit{avang latika} (clove-flavored sweet pastry), \textit{kachori sabzi} (fried stuffed bread with vegetable curry), \textit{tamatar chat} (spiced tomato-based street snack).
    \item \textbf{Public Toilets:} Details about well-maintained public toilet facilities (e.g., the Sulabh complex) in key tourist areas. 
    \item \textbf{General Inquiries:} A broad category addressing common questions about weather, best travel seasons, local transportation, accommodation, selfie spots, and boating facilities. 
\end{itemize}

\begin{figure}
    \centering
    \includegraphics[width=\linewidth]{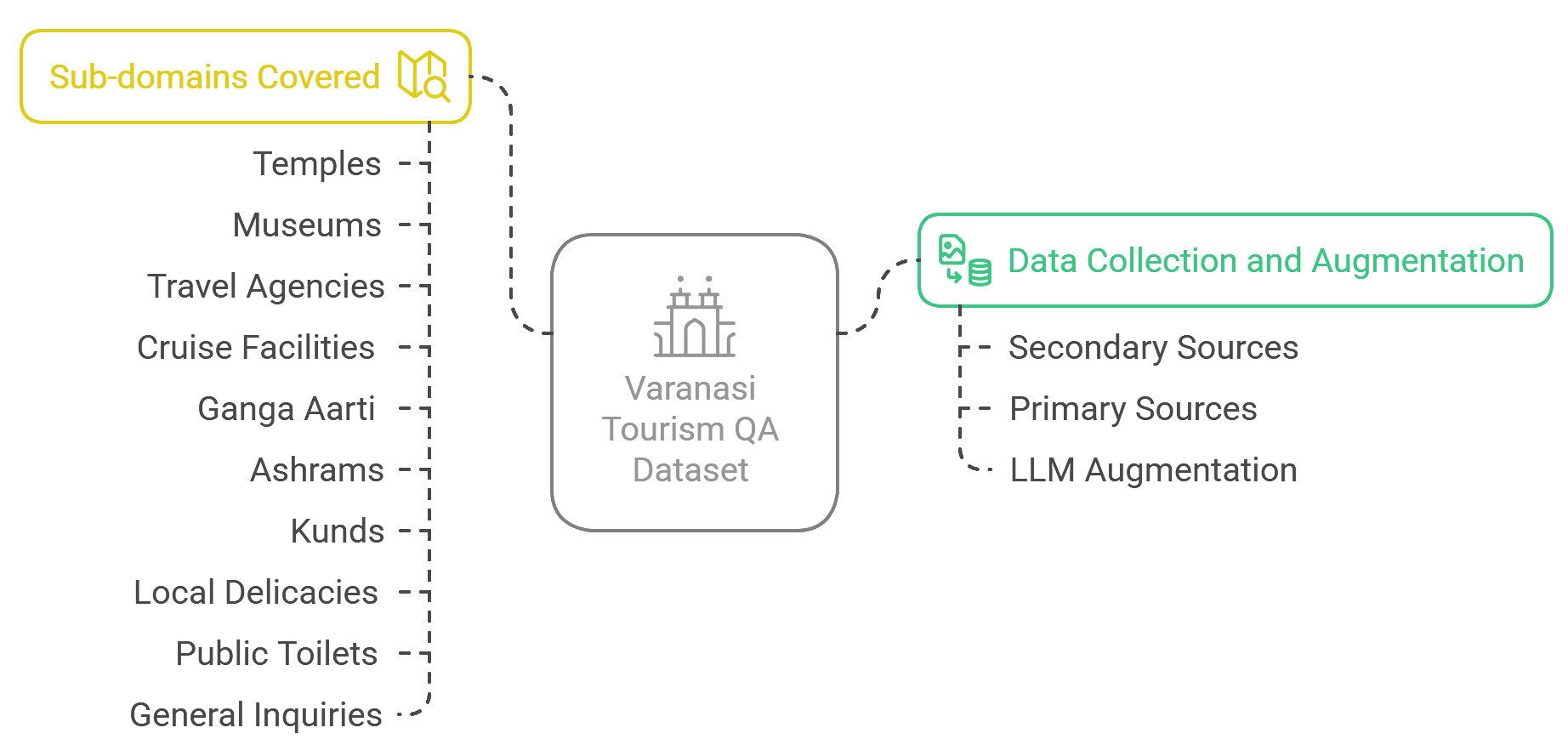}
    \caption{Varanasi Tourism QA Dataset: Structure}
    \label{fig:data_collection}
\end{figure}

To curate this dataset, both purposive and convenience sampling techniques were employed. The outline of data collection, domains, and augmentation is represented in Figure~\ref{fig:data_collection}.

\subsubsection{Purposive Sampling} We selectively targeted prominent and high-demand locations such as: \begin{itemize} \item \textit{Temples:} Including Shri Kashi Vishwanath Mandir, Annapurna Mandir, Sankat Mochan Mandir, Vishalakshi Mandir, etc. \item \textit{Ashrams:} Such as Annapurna Ashram, Kumarswami Ashram, and Jangamwadi Mutt. \item \textit{Travel Agencies:} Including Shreenath Ji Tours and Travels, Endeman Tour, and Ganga Travels. \item \textit{Kunds:} Like Lolark Kund, Durga Kund, Krim Kund, and Manikarnika Kund. \item \textit{Major Activities:} Such as Ganga Aarti and cruise services. \end{itemize}

\subsubsection{Convenience Sampling} Due to constraints such as high location density, limited access to some informants, and time restrictions, convenience sampling was applied to include accessible locations like \textit{Mani Mandir, Rudra Kund, and Udupi 2 Mumbai Food Court}. This ensured that the dataset remained diverse yet practical.

\subsection{Sub-domain-wise Statistics}
The statistics of the data set for various subdomains related to Varanasi tourism are presented, with a focus on the Manually Created Hindi Question Answer Dataset (MCHQAD) and the Llama Generated Hindi Question Answer Dataset (LGHQAD), which was augmented by zero-shot prompting~\cite{scius2025zero}. Among the covered domains, \textit{Temples} constitute the largest category, comprising 2,686 MCHQAD and 9,496 LGHQAD. The \textit{Kunds} subdomain includes 470 MCHQAD and 2,398 LGHQAD, while the \textit{Ashram} category consists of 1,555 MCHQAD and 5,284 LGHQAD. The \textit{Museum} domain includes 484 MCHQAD but does not contain any LGHQAD. The \textit{Travel} The agency domain represents the second largest category in the dataset, with 2,413 MCHQAD and 6,828 LGHQAD. This highlights the significant focus on tourism-related services and infrastructure. 

Additionally, several smaller yet essential subdomains have been included to enhance the comprehensiveness of the dataset. The \textit{Ganga Aarti} category comprises 15 MCHQAD and 46 LGHQAD, reflecting the importance of this spiritual event for visitors. \textit{Cruise Services}, which cater to river tourism experiences, contain 19 MCHQAD and 60 LGHQAD. The \textit{Food Courts} sub-domain, which provides insights into local dining options, includes 11 MCHQAD and 15 LGHQAD. In addition, the public infrastructure is also addressed within the data set. The \textit{Public Toilets} domain includes 9 MCHQAD and 13 LGHQAD, ensuring that accessibility-related queries are covered. The \textit{General Enquiries} category, with 53 MCHQAD and 81 LGHQAD, serves as a broad repository for common tourist questions and essential information. An example QA pair from the dataset is shown in Figure~\ref{fig:example}. The English translation of the example is given below: 
\begin{figure}
    \centering
    \includegraphics[width=0.7\linewidth]{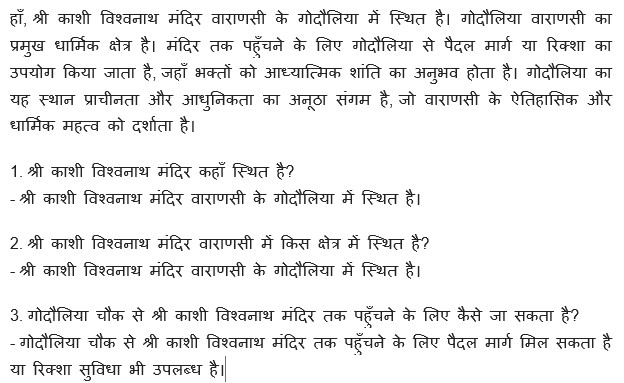}
    \caption{Tourism Domain Example}
    \label{fig:example}
\end{figure}

\textit{``Yes, Shri Kashi Vishwanath Temple is located in Godowlia, Varanasi. Godowlia is the main religious area of Varanasi. Devotees can walk on foot or use a rickshaw from Godowlia to reach the temple, where they can experience spiritual peace. Godowlia is a unique place with the confluence of antiquity and modernity, which reflects the historical and religious importance of Varanasi.\\
\hspace{1 cm} Que. - 1. Where is Shri Kashi Vishwanath Temple located? \\
\hspace{1 cm} Ans. - Shri Kashi Vishwanath Temple is located in Godaulia, Varanasi.\\
\hspace{1 cm} Que. - 2. In which area the Shri Kashi Vishwanath Temple is located in Varanasi?\\
\hspace{1 cm} Ans. - Shri Kashi Vishwanath Temple is located in Godowlia, Varanasi.\\
\hspace{1 cm} Que. - 3. How to reach Shri Kashi Vishwanath Temple from Godaulia Chowk?\\
\hspace{1 cm} Ans. - To reach Shri Kashi Vishwanath Temple from Godowlia Chowk, one can go by foot, or a rickshaw facility is also available."}

These diverse categories collectively contribute to the building of a well-rounded QA system for Varanasi tourism, ensuring coverage of both the main and minor aspects of the tourist experience. The total numbers of MCHQAD and LGHQAD are 7715 and 27455, respectively, as shown in Table~\ref{table:dataset_descrition}. 
A detailed questionnaire was first developed to capture all necessary information related to Varanasi tourism. Data were collected from both secondary sources (as mentioned in Table 14) and primary sources (official temple websites, pamphlets, visiting cards, etc). Primary data were gathered during 10 field visits to key sites, including temples, travel agencies, museums, ashrams, and kunds. The primary data served to validate and refine the information obtained from secondary sources, with rigorous proofreading ensuring accuracy. A total of six annotators were involved in the creation of the dataset. Two annotators created the initial manual dataset, and four were involved in the data augmentation process. Initially, two Hindi language experts validated the manually created dataset, which was revised and proofread multiple times during its development. After the manual creation, the dataset was augmented using the LLaMA model to generate additional similar Hindi questions, which were then cross-checked and manually validated.

\begin{table}[!ht]
\centering
\caption{\textbf{Sub-domain-wise Distribution of QA Dataset for Varanasi Tourism}}
\label{table:dataset_descrition}
\setlength{\tabcolsep}{3pt}
\begin{tabular}{|l|l|l|}
\hline
\bf{Sub-domain} & \bf{\# of MCHQAD} & \bf{\# of LGHQAD} \\
\hline
Temples & 2686 & 11691 \\ \hline
Kunds & 470 & 2398 \\ \hline
Ashrams & 1555 & 5284 \\ \hline
Museums & 484 & 1039 \\ \hline
Travel Agencies & 2413 & 6828 \\ \hline
Ganga  Aarti & 15 & 46 \\ \hline
Cruise & 19 & 60 \\ \hline
Food Court & 11 & 15 \\ \hline
Public Toilet & 9 & 13 \\ \hline
General Enquiries & 53 & 81 \\ \hline
\bf{Total} & \bf{7715} & \bf{27455} \\ \hline
\end{tabular}
\end{table}

\section{Problem Formulation}
\label{sec:problem_statement}
\begin{figure*}
    \centering
    \includegraphics[width=\linewidth]{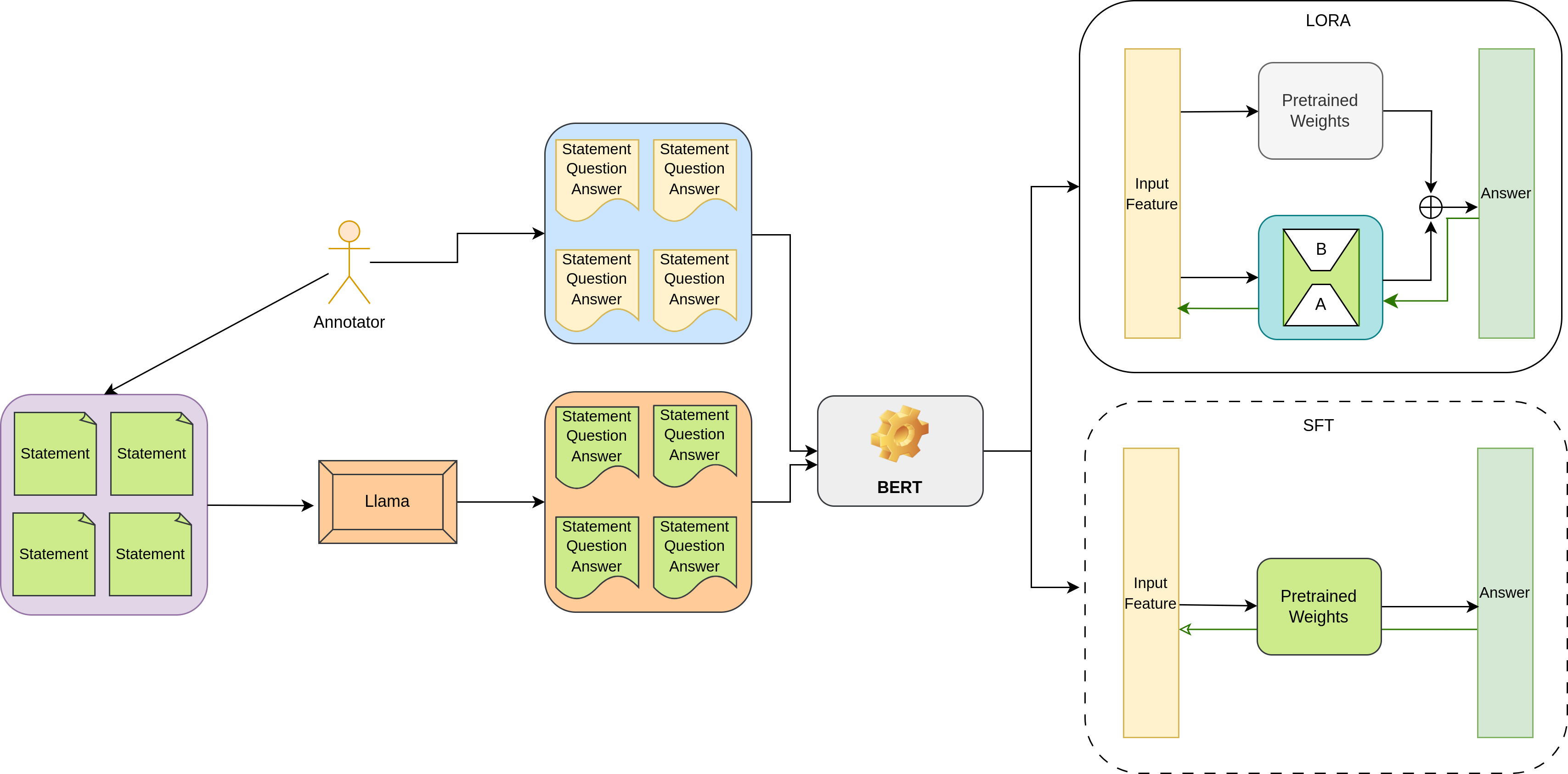}
    \caption{Overview of the complete methodology of the model. The green arrows in the SFT and LoRA modules indicate the back-propagation steps during fine-tuning. At any given time, either SFT or LoRA is activated for modeling the question-answering task.}
    \label{fig:overview}
\end{figure*}

Extractive QA is a fundamental task in NLP that requires models to extract relevant information from a given context to answer a question. Formally, a QA system can be defined as a function:

\begin{equation}
    f: (Q, C) \to A
\end{equation}

where \( Q \) represents the question, \( C \) denotes the context or passage containing the answer, and \( A \) is the extracted answer span. In this article, we relied on Transformer-based modern approaches, such as BERT (Bidirectional Encoder Representations from Transformers), RoBERTa (Robustly Optimized BERT Pretraining Approach), along with associated fine-tuning techniques like LoRA (Low-Rank Adaptation) to achieve high accuracy while maintaining efficiency. The overview of the complete architecture is shown in Figure~\ref{fig:overview}.

\section{Question Answering with BERT}
BERT~\cite{devlin2019bert} is a transformer-based model for contextualized word representations using bidirectional self-attention. Given an input sequence \( X \) of tokens \([ \texttt{[CLS]}, x_1, \ldots, x_n, \texttt{[SEP]} ]\), BERT applies multiple layers of a transformer encoder \( \text{Enc}_{\theta} \), producing contextualized embeddings:
\begin{equation}
    H = \text{Enc}_{\theta}(X) = \{ h_1, h_2 \ldots, h_n \}, \quad h_i \in \mathbb{R}^d
    \label{label:enc_eq}
\end{equation}

BERT pretraining involves Masked Language Modeling (MLM) and Next Sentence Prediction (NSP), where MLM randomly masks tokens \( X_m \subset X \) and predicts them using adjacent words:

\begin{equation}
    L_{\text{MLM}} = - \sum_{x_i \in X_m} \log P_{\theta} (x_i | X_{\text{masked}})
\end{equation}

Similarly, NSP tells the consecutiveness of two sentences \( (S_1, S_2) \) by using:
\begin{equation}
    L_{\text{NSP}} = - y \log P_{\theta} (y | S_1, S_2) - (1 - y) \log (1 - P_{\theta} (y | S_1, S_2))
\end{equation}

Since both MLM and NSP are performed during pretraining, the objective is to minimize the total loss:
\begin{equation}
    L_{\text{BERT}} = L_{\text{MLM}} + L_{\text{NSP}}
\end{equation}

\subsection{Question Answering}
As specified in Section~\ref{sec:problem_statement} for QA input and output, the input sequence to BERT consists of the given question $Q$ and context $C$, represented as:

\begin{equation}
X = [\texttt{[CLS]}; Q; \texttt{[SEP]}; C; \texttt{[SEP]}]
\label{eq:inputformat}
\end{equation}

where \( \texttt{[CLS]} \) is a special classification token and \( \texttt{[SEP]} \) marks boundaries between the question and the context. The model applies multiple layers of a Transformer-based encoder \( \text{Enc}_{\theta} \) (parameterized by \( \theta \)) to generate contextualized token embeddings using equation~\ref{label:enc_eq}

\section{Question Answering With RoBERTa}
RoBERTa~\cite{liu2019roberta} builds upon BERT but introduces modifications to the pretraining and fine-tuning mechanisms, leading to superior performance on QA tasks. Let \( X \) be a sequence of tokens designed using equation~\ref{eq:inputformat} and let \( M(X) \) denote a random masked version of \( X \). In BERT, static masking is applied once, whereas RoBERTa uses dynamic masking, where each token \( x_i \) is masked with a probability \( p_m \) in each batch:

\begin{equation}
X_{\text{masked}} = M(X), \quad M(X) = \{ x_i \text{ with probability } p_m \}
\end{equation}

The MLM loss is computed as:

\begin{equation}
L_{\text{MLM}} = - \sum_{x_i \in M(X)} \log P_{\phi} (x_i | X_{\text{masked}})
\end{equation}

where \( P_{\phi} \) is the probability assigned by RoBERTa’s encoder \( \text{Enc}_{\phi} \) (as shown in equation~\ref{label:enc_eq} parameterized by \( \phi \)).

\section{Supervised Fine-Tuning with Task-Specific Layers}
Supervised fine-tuning the pretrained BERT or RoBERTa involves modifying their architecture by adding task-specific layers and optimizing the model on a downstream task loss function. In the context of extractive QA, this requires training the model to predict the start and end indices of the answer span within a given context.

For span prediction, both models employ two linear classifiers to determine the start and end positions of the answer:

\begin{equation}
P(s) = \text{softmax}(H W_s), \quad P(e) = \text{softmax}(H W_e)
\end{equation}

where \( W_s, W_e \in \mathbb{R}^{d} \) are learned weight vectors for predicting start and end positions, and softmax ensures probabilities sum to 1 across possible token positions. $H$ is the pretrained weights of the models. The model is fine-tuned by minimizing the negative log-likelihood loss for the correct start and end indices \( (s^*, e^*) \):

\begin{equation}
L_{\text{QA}} = -\log P(s = s^*) - \log P(e = e^*)
\end{equation}

where \( s^* \) and \( e^* \) are the starting and end positions of the ground truth, while $s$ and $e$ are respective positions obtained from prediction. The model maximizes the probability of selecting the correct answer span.

\subsection{Fine-Tuning with LoRA}
Fine-tuning BERT or RoBERTa on large QA datasets requires updating all parameters \( \theta \), leading to high computational costs. The standard parameter updation expression is represented by:

\begin{equation}
\theta_{\text{new}} = \theta - \eta \nabla_{\theta} L_{\text{QA}}
\end{equation}

where \( \eta \) is the learning rate. This full fine-tuning is expensive.
To address this, LoRA (Low-Rank Adaptation)~\cite{hu2022lora} freezes the original model parameters and injects low-rank updates into weight matrices. Instead of modifying a full weight matrix \( W \in \mathbb{R}^{d \times k} \), LoRA decomposes updates as:

\begin{equation}
\Delta W = B A^\top
\end{equation}

where \( B \in \mathbb{R}^{d \times r} \), \( A \in \mathbb{R}^{k \times r} \) are low-rank matrices and \( r \ll \min(d, k) \) ensures a small number of trainable parameters. The adapted representation for layer \( l \) becomes:

\begin{equation}
H^l = (W^l + \Delta W^l) H^{l-1}
\end{equation}

To maintain stability, the loss function for LoRA fine-tuning regularizes low-rank updates:

\begin{equation}
L_{\text{LoRA}} = L_{\text{QA}} + \lambda \| B A^\top \|_F^2
\end{equation}

where \( \| \cdot \|_F \) is the Frobenius norm, and \( \lambda \) is a hyperparameter controlling regularization. The algorithm~\ref{alg:qa_lora} describes the SFT and LORA-based fine-tuning. 

\begin{algorithm}[!t]
\small
\SetAlgoLined
\DontPrintSemicolon
\SetKwInput{KwIn}{Input}
\SetKwInput{KwOut}{Output}
\SetKwComment{Comment}{$\triangleright$\ }{}

\KwIn{Pretrained encoder $\mathrm{Enc}_{\theta}$ (BERT/RoBERTa), QA dataset $\mathcal{D} = \{(Q_i, C_i, A_i)\}_{i=1}^N$, learning rate $\eta$, batch size $B$, adapter rank $r$, regularization weight $\lambda$}
\KwOut{Fine-tuned weights $\theta_{\text{SFT}}$ and or LORA adapter $\theta_{\text{LORA}}$}

\Comment*[l]{\textbf{Preprocessing}}
\For{$i \gets 1$ \KwTo $N$}{
    $X_i \gets [\texttt{[CLS]}; Q_i; \texttt{[SEP]}; C_i; \texttt{[SEP]}]$ \tcp*{Tokenize input}
    $(s^*_i, e^*_i) \gets \text{align}(X_i, A_i)$ \tcp*{Map answer to token indices}
}

\BlankLine
\Comment*[l]{\textbf{SFT}}
\ForEach{batch $\mathcal{B} \subset \mathcal{D}$ \textbf{with} $|\mathcal{B}| = B$}{
    $L_{\text{batch}} \gets 0$ \;
    \ForEach{$(Q, C, s^*, e^*) \in \mathcal{B}$}{
        $X \gets [\texttt{[CLS]}; Q; \texttt{[SEP]}; C; \texttt{[SEP]}]$ \;
        $H \gets \mathrm{Enc}_{\theta}(X)$ \tcp*{Contextual embedding}
        $P_s \gets \text{softmax}(H W_s)$; $P_e \gets \text{softmax}(H W_e)$ \tcp*{Start/end probs}
        $L_{\text{QA}} \gets -\log P_s[s^*] - \log P_e[e^*]$ \tcp*{QA loss}
        $L_{\text{batch}} \gets L_{\text{batch}} + L_{\text{QA}}$ \;
    }
    $\theta \gets \theta - \eta \nabla_{\theta} L_{\text{batch}}$ \tcp*{Gradient update}
}
\Return{\(\theta_{\text{SFT}}\)} \;
\BlankLine
\Comment*[l]{\textbf{LORA}}
Freeze $\theta$; initialize $A^l \in \mathbb{R}^{k \times r}$, $B^l \in \mathbb{R}^{d \times r}$ \tcp*{Adapter params}
\ForEach{batch $\mathcal{B} \subset \mathcal{D}$}{
    $L_{\text{batch}} \gets 0$ \;
    \ForEach{$(Q, C, s^*, e^*) \in \mathcal{B}$}{
        $X \gets [\texttt{[CLS]}; Q; \texttt{[SEP]}; C; \texttt{[SEP]}]$ \;
        $W^l_{\text{eff}} \gets W^l + B^l (A^l)^\top$ \tcp*{LoRA update}
        $H \gets \mathrm{Enc}_{\theta}(X; W^l_{\text{eff}})$ \tcp*{Adapted encoding}
        Compute $P_s$, $P_e$, $L_{\text{QA}}$ as in SFT \;
        $L_{\text{batch}} \gets L_{\text{batch}} + L_{\text{QA}}$ \;
    }
    $L_{\text{reg}} \gets \sum_l \| B^l (A^l)^\top \|_F^2$ \tcp*{Frobenius regularization}
    $L \gets L_{\text{batch}} + \lambda L_{\text{reg}}$ \;
    $(A^l, B^l) \gets (A^l, B^l) - \eta \nabla_{A^l, B^l} L$ \tcp*{Update adapters}
}
\Return{\(\theta_{\text{LORA}} \gets \{A^l, B^l\}\)} \;

\caption{\textbf{Extractive QA Fine-Tuning with BERT/RoBERTa using SFT and LoRA}}
\label{alg:qa_lora}
\end{algorithm}

\section{Experimental Settings}
For preprocessing, the input Hindi QA data was tokenized with a maximum sequence length of 384 tokens and a stride of 128 tokens to effectively capture overlapping contexts; additionally, both overflowing tokens and offset mappings were returned, and all sequences were padded to the maximum length to ensure uniformity. 
We conduct an extensive evaluation of two transformer-based language models, multilingual foundation Models-BERT and RoBERTa, employing SFT on a multi-domain dataset comprising diverse sub-domains such as temples, museums, travel agencies, cruise facilities, and other culturally significant areas. We have used \texttt{bert-base-multilingual-cased}~\cite{devlin2019bert}, \texttt{ai4bharat/indic-bert}~\cite{kakwani2020indicnlpsuite}, \texttt{l3cube-pune/hindi\\-bert-v2}~\cite{joshi2022l3cube} and \texttt{l3cube-pune/hindi-roberta}~\cite{joshi2022l3cube}, referred as \textbf{mBERT}, \textbf{IndicBERT}, \textbf{HindiBERT} and \textbf{Hindi-RoBERTa}, respectively. The dataset has been split into 80 and 20 rations for training and testing, respectively.  Both models were fine-tuned under a consistent hyperparameter regime, using a learning rate of 3e-5, a batch size of 48, and training for upto 3 epochs with early stopping based on validation loss, while optimization was carried out using the AdamW optimizer. To further promote regularization, a weight decay of 0.01 was also employed. We targeted the model's query and value modules for adaptation, applying a dropout rate of 0.1 to the LoRA layers and opting for no bias adaptation, thereby ensuring parameter efficiency and robust model adjustment. The LoRA exclusively used with mBERT by introducing trainable low-rank matrices. We have explored LoRA configurations with rank values of 2, 4, 8, 16, and 32.

For data augmentation, the Llama model has explored with zero-shot propmpting~\cite{scius2025zero} where the prompt was \texttt{Statement: Generate questions in Hindi along with their answers from \textit{Context}}.

\section{Results and Discussion}

\begin{table}[!hbt]
\caption{\textbf{Ganga Aarti Domain Model Scores}}
\label{table:Aarti}
\label{table}
    \centering
    \setlength{\tabcolsep}{3pt}
    \begin{tabular}{|p{60pt}|p{40pt}|p{35pt}|p{35pt}|p{35pt}|p{35pt}|}
    \hline
    \textbf{Model} & \textbf{F1 Score} & \textbf{BLEU} & \textbf{RougeL} \\ \hline
IndicBERT &  8.67 & 2.361 & 20\\ \hline
BERT &  52.664 & 28.578 & 20\\ \hline
LORA-bert(r2) &  13.979 & 18.203 & 0\\ \hline
LORA-bert(r4) &  16.512 & 7.477 & 0\\ \hline
LORA-bert(r8) &  15.569 & 6.409 & 0\\ \hline
LORA-bert(r16) &  15.569 & 6.409 & 0\\ \hline
LORA-bert(r32) &  15.569 & 6.409 & 0\\ \hline
HindiBERT &  57.774 & 29.748 & 20\\ \hline
Hindi-RoBERTa &  63.064 & 45.539 & 20\\ \hline
\end{tabular}
\end{table}

Table~\ref{table:Aarti}, for the Aarti domain, the F1, BLEU, and RougeL scores for the IndicBERT model were found to be 8.67, 2.361, and 20 respectively. The BERT model had an F1 Score of 52.664, a BLEU score of \& 28.578, and RougeL score of 20. In the case of the LORA-bert(r2) model, the F1 Score was 13.979, BLEU score was 8.203, and RougeL score was 0. The LORA-bert(r4) model had an F1 Score of 16.512, BLEU score of 7.477, and RougeL score was found to be 0. The LORA-bert(r8) model had an F1 Score of 15.569, BLEU score of 6.409, and RougeL score was found to be 0. The LORA-bert(r16) model had an F1 Score of 15.569, BLEU score of 6.409, and RougeL score was 0. The LORA-bert(r32) model had an F1 Score of 15.569, BLEU score of 6.409, and RougeL score was 0. For the HindiBERT model, F1 Score was found to be 57.774, BLEU score 29.748, and RougeL score was 20. The F1, BLEU, and RougeL scores for the Hindi-RoBERTa model, were found to be 63.064, 45.539 and 20, respectively.

\begin{table}[!ht]
\caption{\textbf{Cruise Domain Model Score}}
\label{table:Cruise}
\label{table}
    \centering
    \setlength{\tabcolsep}{3pt}
    \begin{tabular}{|p{60pt}|p{40pt}|p{35pt}|p{35pt}|p{35pt}|p{35pt}|}
    \hline
    \textbf{Model} & \textbf{F1 Score} & \textbf{BLEU} & \textbf{RougeL} \\ \hline
IndicBERT & 6.691 & 0.875 & 33.333\\ \hline
BERT & 73.477 & 50.658 & 34.999\\ \hline
LORA-bert(r2) & 12.267 & 3.032 & 0\\ \hline
LORA-bert(r4) & 12.05 & 3.829 & 0\\ \hline
LORA-bert(r8) & 12.267 & 3.032 & 0\\ \hline
LORA-bert(r16) & 12.267 & 3.032 & 0\\ \hline
LORA-bert(r32) & 12.267 & 3.032 & 0\\ \hline
HindiBERT & 30.02 & 15.925 & 45\\ \hline
Hindi-RoBERTa & 41.843 & 20.305 & 35\\ \hline
\end{tabular}
\end{table}

Table~\ref{table:Cruise}, for the Cruise domain, the F1, BLEU, and RougeL scores of the IndicBERT model were found to be 6.691, 0.875 and 33.333 respectively. The BERT model had an F1 Score of 73.477, BLEU score was 50.658, and RougeL score 34.999. Whereas in case of the LORA-bert(r2) model, the F1 Score was 12.267, BLEU score was 3.032, and RougeL score was 0. The LORA-bert(r4) model had an F1 Score of 12.05, BLEU score was 3.829, and RougeL score was 0. The LORA-bert(r8) model had an F1 Score of 12.267, BLEU score was 3.032, and RougeL score was 0. The LORA-bert(r16) model had an F1 Score of 12.267, BLEU score was 3.032, and RougeL score was 0. The LORA-bert(r32) model had an F1 Score of 13.605, BLEU score was 9.7, and RougeL score was 0. HindiBERT had an F1 Score of 30.02, BLEU score was 15.925, and RougeL score was 45. The F1, BLEU, and RougeL scores for the Hindi-RoBERTa model were found to be 41.843, 20.305, and 35, respectively.

\begin{table}[!ht]
\caption{\textbf{Food Court Domain Model Scores}}
\label{table:Food Court}
\label{table}
    \centering
    \setlength{\tabcolsep}{3pt}
    \begin{tabular}{|p{60pt}|p{40pt}|p{35pt}|p{35pt}|p{35pt}|p{35pt}|}
    \hline
    \textbf{Model} & \textbf{F1 Score} & \textbf{BLEU} & \textbf{RougeL} \\ \hline
IndicBERT & 1.666 & 0.992 & 0\\ \hline
BERT & 53.333 & 34.042 & 0\\ \hline
LORA-bert(r2) & 2.469 & 0.511 & 0\\ \hline
LORA-bert(r4) & 2.469 & 0.511 & 0\\ \hline
LORA-bert(r8) & 2.469 & 0.511 & 0\\ \hline
LORA-bert(r16) & 2.469 & 0.511 & 0\\ \hline
LORA-bert(r32) & 2.469 & 0.511 & 0\\ \hline
HindiBERT & 52.197 & 32.837 & 0\\ \hline
Hindi-RoBERTa & 69.334 & 46.16 & 0\\ \hline
\end{tabular}
\end{table}

Table~\ref{table:Food Court}, for the Food Court domain, the F1, BLEU, and RougeL scores for the IndicBERT model were found to be 1.666, 0.992, and 0 respectively. The BERT model had the F1 Score of 53.333, BLEU score was 34.042, and RougeL score was 0. Whereas in case of the LORA-bert(r2) model, the F1 Score was 2.469, BLEU score was 0.511, and RougeL score was 0. The LORA-bert(r4) model had an F1 Score of 2.469, BLEU score was 0.511, and RougeL score was 0. The LORA-bert(r8) model had an F1 Score of 2.469, BLEU score was 0.511, and RougeL score was 0. The LORA-bert(r16) model had an F1 Score of 2.469, BLEU score was 0.511, and RougeL score was 0. The LORA-bert(r32) model had an F1 Score of 2.469, BLEU score was 0.511, and RougeL score was 0. HindiBERT had an F1 Score of 52.197, BLEU score was 32.837, and RougeL was score 0. The F1, BLEU, and RougeL scores for the Hindi-RoBERTa model were found to be 69.334, 46.16, and 0, respectively.

\begin{table}[!ht]
\caption{\textbf{Toilet Domain Model Scores}}
\label{table:Toilet}
\label{table}
    \centering
    \setlength{\tabcolsep}{3pt}
    \begin{tabular}{|p{60pt}|p{40pt}|p{35pt}|p{35pt}|p{35pt}|p{35pt}|}
    \hline
    \textbf{Model} & \textbf{F1 Score} & \textbf{BLEU} & \textbf{RougeL} \\ \hline
IndicBERT & 0 & 0 & 0\\ \hline
BERT & 73.469 & 30.786 & 66.666\\ \hline
LORA-bert(r2) & 33.939 & 31.252 & 0\\ \hline
LORA-bert(r4) & 33.939 & 31.252 & 0\\ \hline
LORA-bert(r8) & 33.939 & 31.252 & 0\\ \hline
LORA-bert(r16) & 33.939 & 31.252 & 0\\ \hline
LORA-bert(r32) & 33.939 & 31.252 & 0\\ \hline
HindiBERT & 64.285 & 58.456 & 66.666\\ \hline
Hindi-RoBERTa & 66.666 & 64.118 & 66.666\\ \hline
\end{tabular}
\end{table}
Table~\ref{table:Toilet}, for the Toilet domain, the F1, BLEU, and RougeL scores for the IndicBERT model were found to be 0, 0, and 0 respectively. The BERT model had an F1 Score of 73.469, BLEU score was 30.786, and RougeL score was 66.666. Whereas in case of the LORA-bert(r2) model, the F1 Score was 33.939, BLEU score was 31.252, and RougeL score was 0. The LORA-bert(r4) model had an F1 Score of 33.939, BLEU score was 31.252, and RougeL score was 0. The LORA-bert(r8) model had an F1 Score of 33.939, BLEU score was 31.252, and RougeL score was 0. The LORA-bert(r16) model had an F1 Score of 33.939, BLEU score was 31.252, and RougeL score was 0. The LORA-bert(r32) model had an F1 Score of 33.939, BLEU score was 31.252, and RougeL score was 0. HindiBERT had an F1 Score of 64.285, BLEU score was 58.456, and RougeL score was 66.666. The F1, BLEU, and RougeL scores for the Hindi-RoBERTa model were found to be 66.666, 64.118, and 66.666, respectively.

\begin{table}[!ht]
\caption{\textbf{Kund Domain Model Scores}}
\label{table:Kund}
\label{table}
    \centering
    \setlength{\tabcolsep}{3pt}
    \begin{tabular}{|p{60pt}|p{40pt}|p{35pt}|p{35pt}|p{35pt}|p{35pt}|}
    \hline
    \textbf{Model} & \textbf{F1 Score} & \textbf{BLEU} & \textbf{RougeL} \\ \hline
IndicBERT & 5.357 & 0.352 & 14.374\\ \hline
BERT & 72.14 & 64.559 & 14.791\\ \hline
LORA-bert(r2) & 6.312 & 1.99 & 0.208\\ \hline
LORA-bert(r4) & 11.258 & 10.105 & 0.208\\ \hline
LORA-bert(r8) & 7.063 & 3.868 & 0\\ \hline
LORA-bert(r16) & 7.506 & 4.231 & 0.416\\ \hline
LORA-bert(r32) & 9.919 & 5.314 & 0.416\\ \hline
HindiBERT & 59.995 & 42.809 & 13.958\\ \hline
Hindi-RoBERTa & 69.207 & 56.497 & 14.583\\ \hline
\end{tabular}
\end{table}

Table~\ref{table:Kund}, for the Kund domain, the F1, BLEU, and RougeL scores for the IndicBERT model were found to be 5.357, 0.352 and 14.374 respectively. The BERT model had an F1 Score of 72.14, BLEU score was 64.559, and RougeL score was 14.791. Whereas in case of the LORA-bert(r2) model, the F1 Score was 6.312, BLEU score was 1.99, and RougeL score was 0.208. The LORA-bert(r4) model had an F1 Score of 11.258, BLEU score was 10.105, and RougeL score was 0.208. The LORA-bert(r8) model had an F1 Score of 7.063, BLEU score was 3.868, and RougeL score was 0. The LORA-bert(r16) model had an F1 Score of 7.506, BLEU score was 4.231, and RougeL score was 0.416. The LORA-bert(r32) model had an F1 Score of 9.919, BLEU score was 5.314, and RougeL score was  0.416. HindiBERT had an F1 Score of 59.995, BLEU score was 42.809, and RougeL score was 13.958. The F1, BLEU, and RougeL scores for the Hindi-RoBERTa model were found to be 69.207, 56.497 and 14.583 respectively.

\begin{table}[!ht]
\caption{\textbf{Museum Domain Model Scores}}
\label{table:Museum}
\label{table}
    \centering
    \setlength{\tabcolsep}{3pt}
    \begin{tabular}{|p{60pt}|p{40pt}|p{35pt}|p{35pt}|p{35pt}|p{35pt}|}
    \hline
    \textbf{Model} & \textbf{F1 Score} & \textbf{BLEU} & \textbf{RougeL} \\ \hline
IndicBERT & 7.998 & 4.089 & 39.523\\ \hline
BERT & 92.738 & 85.851 & 32.743\\ \hline
LORA-bert(r2) & 57.118 & 48.933 & 24.007\\ \hline
LORA-bert(r4) & 68.093 & 59.078 & 26.984\\ \hline
LORA-bert(r8) & 65.344 & 57.676 & 26.825\\ \hline
LORA-bert(r16) & 65.965 & 59.367 & 27.38\\ \hline
LORA-bert(r32) & 67.663 & 59.484 & 27.936\\ \hline
HindiBERT & 91.71 & 80.869 & 39.523\\ \hline
Hindi-RoBERTa & 98.887 & 97.674 & 39.523\\ \hline
\end{tabular}
\end{table}

Table~\ref{table:Museum}, for the Museum domain, the F1, BLEU, and RougeL scores for the IndicBERT model were found to be 7.998, 4.089, and 39.523 respectively. The BERT model had an F1 Score of 92.283, BLEU score was 85.851, and RougeL score was 32.743. Whereas in case of the LORA-bert(r2) model, the F1 Score was 57.118, BLEU score was 48.933, and RougeL score was 24.007. The LORA-bert(r4) model had an F1 Score of 968.093, BLEU score  was 59.078, and RougeL score was 26.984. The LORA-bert(r8) model had an F1 Score of 65.344, BLEU score was 57.676, and RougeL was score 26.825. The LORA-bert(r16) model had an F1 Score of 65.965, BLEU score was 59.367, and RougeL score was 27.38. The LORA-bert(r32) model had an F1 Score of 67.663, BLEU score was 59.484, and RougeL score was 27.936. HindiBERT had an F1 Score of 91.71, BLEU score was 80.869, and RougeL score was 39.523. The F1, BLEU, and RougeL scores for the Hindi-RoBERTa model were found to be 98.887, 97.674, and 39.523 respectively.

\begin{table}[!ht]
\caption{\textbf{General Domain Model Scores}}
\label{table:General}
\label{table}
    \centering
    \setlength{\tabcolsep}{3pt}
    \begin{tabular}{|p{60pt}|p{40pt}|p{35pt}|p{35pt}|p{35pt}|p{35pt}|}
    \hline
    \textbf{Model}  & \textbf{F1 Score} & \textbf{BLEU} & \textbf{RougeL} \\ \hline
IndicBERT & 8.353 & 0.984 & 21.568\\ \hline
BERT & 92.715 & 84.223 & 23.529\\ \hline
LORA-bert(r2) & 19.307 & 17.552 & 10.924\\ \hline
LORA-bert(r4) & 15.142 & 18.916 & 3.921\\ \hline
LORA-bert(r8) & 15.142 & 18.768 & 3.921\\ \hline
LORA-bert(r16) & 15.142 & 18.768 & 3.921\\ \hline
LORA-bert(r32) & 15.142 & 18.916 & 3.921\\ \hline
HindiBERT & 71.698 & 66.071 & 23.529\\ \hline
Hindi-RoBERTa & 73.103 & 74.271 & 23.529\\ \hline
\end{tabular}
\end{table}

Table~\ref{table:General}, for the General domain, the F1, BLEU, and RougeL scores for the IndicBERT model were found to be 8.353, 0.984, and 21.568 respectively. The BERT model had an F1 score of 92.715, the BLEU score was 84.223, and the RougeL score was 23.529. whereas in case of the LORA-bert(r2) model, the F1 score was 19.307, the BLEU score was 17.552, and the RougeL score was 10.924. The LORA-bert(r4) model had an F1 score of 15.142, the BLEU score was 18.916, and the RougeL score was 3.921. The LORA-bert(r8) model had an F1 score of 15.142, the BLEU score was 18.768, and the RougeL score was 3.921. The LORA-bert(r16) model had an F1 score of 15.142, the BLEU score was 18.768, and the RougeL score was 3.921. The LORA-bert(r32) model had an F1 score of 15.142, the BLEU score was 18.916, and the RougeL score was 3.921. HindiBERT had an F1 score of 71.698, the BLEU score was 66.071, and the RougeL score was 23.529. The F1, BLEU, and RougeL scores for the Hindi-RoBERTa model were found to be 73.103, 74.271, and 23.529 respectively.

\begin{table}[!ht]
\caption{\textbf{Ashram Domain Model Scores}}
\label{table:Aashram}
    \centering
    \setlength{\tabcolsep}{3pt}
    \begin{tabular}{|p{60pt}|p{40pt}|p{35pt}|p{35pt}|p{35pt}|p{35pt}|}
    \hline
    \textbf{Model}  & \textbf{F1 Score} & \textbf{BLEU} & \textbf{RougeL} \\ \hline
IndicBERT & 8.469 & 2.909 & 48.085\\ \hline
BERT & 92.74 & 85.889 & 48.459\\ \hline
LORA-bert(r2) & 90.846 & 84.657 & 46.825\\ \hline
LORA-bert(r4) & 90.909 & 84.91 & 46.825\\ \hline
LORA-bert(r8) & 90.908 & 84.74 & 46.778\\ \hline
LORA-bert(r16) & 90.991 & 84.588 & 46.825\\ \hline
LORA-bert(r32) & 90.934 & 84.765 & 46.732\\ \hline
HindiBERT & 91.518 & 84.759 & 47.432\\ \hline
Hindi-RoBERTa & 96.739 & 95.2 & 48.482\\ \hline
\end{tabular}
\end{table}

Table~\ref{table:Aashram}, for the Ashram domain, the F1, BLEU and RougeL scores for the IndicBERT model were found to be 8.469, 2.909 and 48.085, respectively. The BERT model had the F1 Score of 92.74, BLEU score was 85.889, and RougeL score was 48.459 whereas in case of the LORA-bert(r2) model, the F1 Score was 90.846, BLEU score was 84.657, and RougeL score was 46.825. The LORA-bert(r4) model had an F1 Score of 90.909, BLEU score was 84.91, and RougeL score was 46.825. The LORA-bert(r8) model had an F1 Score of 90.908, BLEU score was 84.74, and RougeL score was 46.778. The LORA-bert(r16) model had an F1 Score of 90.991, BLEU score was 84.588, and RougeL score was 46.825. The LORA-bert(r32) model had an F1 Score of 90.934, BLEU score was 84.765, and RougeL score was 46.732. HindiBERT had an F1 Score of 91.518, BLEU score was 84.759, and RougeL score was 47.432. The F1, BLEU, and RougeL scores for the Hindi-RoBERTa model were found to be 96.739, 95.2, 48.482 respectively.

\begin{table}[!ht]
\caption{\textbf{Travel Domain Model Scores}}
\label{table:Travel}
\label{table}
    \centering
    \setlength{\tabcolsep}{3pt}
    \begin{tabular}{|p{60pt}|p{40pt}|p{35pt}|p{35pt}|p{35pt}|p{35pt}|}
    \hline
    \textbf{Model}  & \textbf{F1 Score} & \textbf{BLEU} & \textbf{RougeL} \\ \hline
IndicBERT & 1.299 & 0.081 & 1.437\\ \hline
BERT & 19.569 & 7.564 & 1.437\\ \hline
LORA-bert(r2) & 19.983 & 7.432 & 1.437\\ \hline
LORA-bert(r4) & 19.865 & 7.342 & 1.408\\ \hline
LORA-bert(r8) & 19.989 & 7.534 & 1.437\\ \hline
LORA-bert(r16) & 19.999 & 7.547 & 1.437\\ \hline
LORA-bert(r32) & 19.982 & 7.546 & 1.437\\ \hline
HindiBERT & 88.842 & 79.401 & 14.73\\ \hline
Hindi-RoBERTa & 93.526 & 89.755 & 14.612\\ \hline
\end{tabular}
\end{table}

Table~\ref{table:Travel}, for the Travel domain, the F1, BLEU, and RougeL scores for the IndicBERT model were found to be 1.299, 0.081, and 1.437 respectively. The BERT model had the F1 Score of 19.569, BLEU score 7.564, and RougeL score 1.437. Whereas in case of the LORA-bert(r2) model, the F1 Score was 19.983, BLEU score was 7.432, and RougeL score was 1.437. The LORA-bert(r4) model had an F1 Score of 19.865, BLEU score was 7.342, and RougeL score was 1.437. The LORA-bert(r8) model had an F1 Score of 19.989, BLEU score was 7.534, and RougeL score was 1.437. The LORA-bert(r16) model had an F1 Score of 19.999, BLEU score was 7.547, and RougeL score was 1.437. The LORA-bert(r32) model had an F1 Score of 19.982, BLEU score was 7.546, and RougeL score was 1.437. HindiBERT had an F1 Score of 88.842, BLEU score was 79.401, and RougeL score was 14.73. The F1, BLEU, and RougeL scores for the Hindi-RoBERTa model were found to be 93.526, 89.755, and 14.612 respectively.

\begin{table}[!ht]
\caption{\textbf{Temple Domain Model Scores}}
\label{table:Temple}
\label{table}
    \centering
    \setlength{\tabcolsep}{3pt}
    \begin{tabular}{|p{60pt}|p{40pt}|p{35pt}|p{35pt}|p{35pt}|p{35pt}|}
    \hline
    \textbf{Model} & \textbf{F1 Score} & \textbf{BLEU} & \textbf{RougeL} \\ \hline
IndicBERT & 8.21 & 2.533 & 34.75\\ \hline
BERT & 84.232 & 77.012 & 35.718\\ \hline
LORA-bert(r2) & 75.185 & 64.233 & 34.339\\ \hline
LORA-bert(r4) & 75.442 & 64.958 & 34.312\\ \hline
LORA-bert(r8) & 75.424 & 65.068 & 34.318\\ \hline
LORA-bert(r16) & 75.279 & 64.623 & 34.183\\ \hline
LORA-bert(r32) & 75.636 & 65.75 & 34.201\\ \hline
HindiBERT & 75.562 & 63.918 & 34.187\\ \hline
Hindi-RoBERTa & 86.493 & 81.747 & 35.434\\ \hline
\end{tabular}
\end{table}

Table~\ref{table:Temple}, for the Temple domain, the F1, BLEU, and RougeL scores for the IndicBERT model were found to be 8.21, 2.533, and 34.75 respectively. The BERT model had the F1 Score of 84.232, BLEU score  was 77.012, and RougeL score was 35.718. Whereas in case of the LORA-bert(r2) model, the F1 Score was 75.185, BLEU score was 64.233, and RougeL score was 34.339. The LORA-bert(r4) model had an F1 Score of 75.442, BLEU score was 64.958, and RougeL score was 34.312. The LORA-bert(r8) model had an F1 Score of 75.424, BLEU score was 65.068, and RougeL score was 34.318. The LORA-bert(r16) model had an F1 Score of 75.279, BLEU score was 64.623, and RougeL score was 34.183. The LORA-bert(r32) model had an F1 Score of 75.636, BLEU score was 65.75, and RougeL score was 34.201. HindiBERT had an F1 Score of 75.562, BLEU score was 63.918, and RougeL score was 34.187. The F1, BLEU, and RougeL scores for the Hindi-RoBERTa model were found to be 86.493, 81.747, and 35.434 respectively.

\begin{table}[!ht]
\caption{\textbf{Merged Domain Model Scores}}
\label{table:Merged}
\label{table}
    \centering
    \setlength{\tabcolsep}{3pt}
    \begin{tabular}{|p{60pt}|p{40pt}|p{35pt}|p{35pt}|p{35pt}|p{35pt}|}
    \hline
    \textbf{Model}  & \textbf{F1 Score} & \textbf{BLEU} & \textbf{RougeL} \\ \hline
IndicBERT & 0.644 & 0.043 & 1.073\\ \hline
BERT & 13.825 & 3.057 & 0.779\\ \hline
LORA-bert (r2) & 21.082 & 9.881 & 3.452\\ \hline
LORA-bert (r4) & 20.942 & 9.967 & 3.404\\ \hline
LORA-bert (r8) & 14.209 & 3.069 & 1.136\\ \hline
LORA-bert (r16) & 14.139 & 3.028 & 1.125\\ \hline
LORA-bert (r32) & 14.106 & 3.106 & 1.095\\ \hline
HindiBERT & 65.208 & 49.477 & 17.261\\ \hline
Hindi-RoBERTa & 21.409 & 11.008 & 3.349\\ \hline
\end{tabular}
\end{table}

Table~\ref{table:Merged}, for the Merged domain, the F1, BLEU, and RougeL scores for the IndicBERT model were found to be 0.644, 0.043, and 1.073 respectively. The BERT model had an F1 Score of 13.825, BLEU score was 3.057, and RougeL score was 0.779. Whereas in the case of the LORA-bert(r2) model, the F1 Score was 21.082, BLEU score was 9.881, and RougeL score was 3.452. The LORA-bert(r4) model had an F1 Score of 20.942, BLEU score was 9.967, and RougeL score was 3.404. The LORA-bert(r8) model had an F1 Score of 14.209, BLEU score was 3.069, and RougeL was score 1.136. The LORA-bert(r16) model had an F1 Score of 14.139, BLEU score was 3.028, and RougeL score was 1.125. The LORA-bert(r32) model had an F1 Score of 14.106, BLEU score was 3.106, and RougeL score was 1.095. HindiBERT had an F1 Score of 65.208, BLEU score was 49.477, and RougeL score was 17.261. The F1, BLEU, and RougeL scores for the Hindi-RoBERTa model were found to be 21.409, 11.008, and 3.349, respectively.

\begin{figure}[!hbt]
    \centering
    \includegraphics[width=0.5\linewidth]{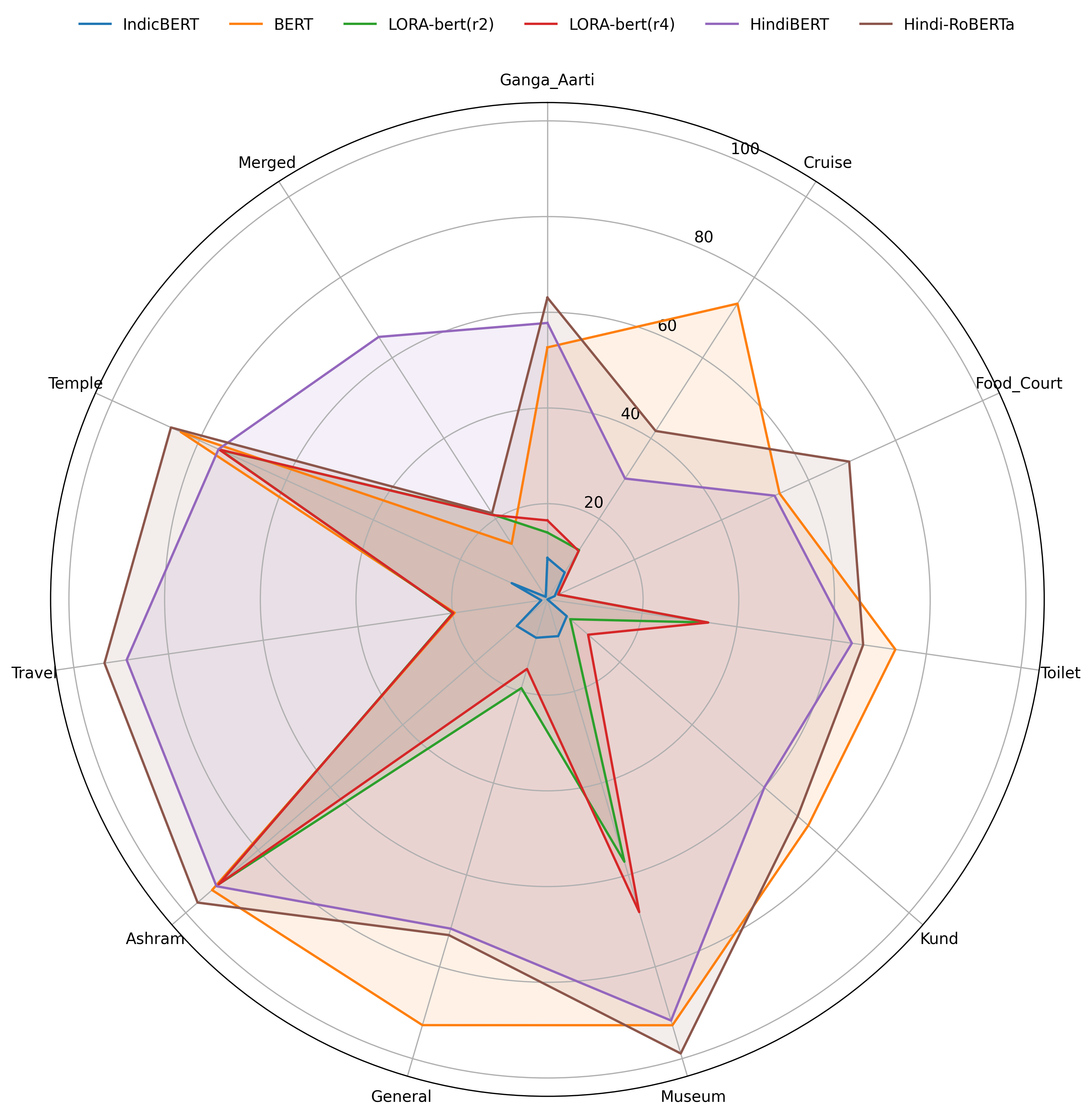}
    \caption{F1 score comparison of the models across 11 domain settings. Configurations with LoRA-based adapters at ranks r8, r16, and r32 consistently underperformed compared to r2 and r4; thus, they are omitted from this figure for clarity.}
    \label{fig:result_compare}
\end{figure}

\begin{figure}[!hbt]
    \centering
    \includegraphics[width=0.5\linewidth]{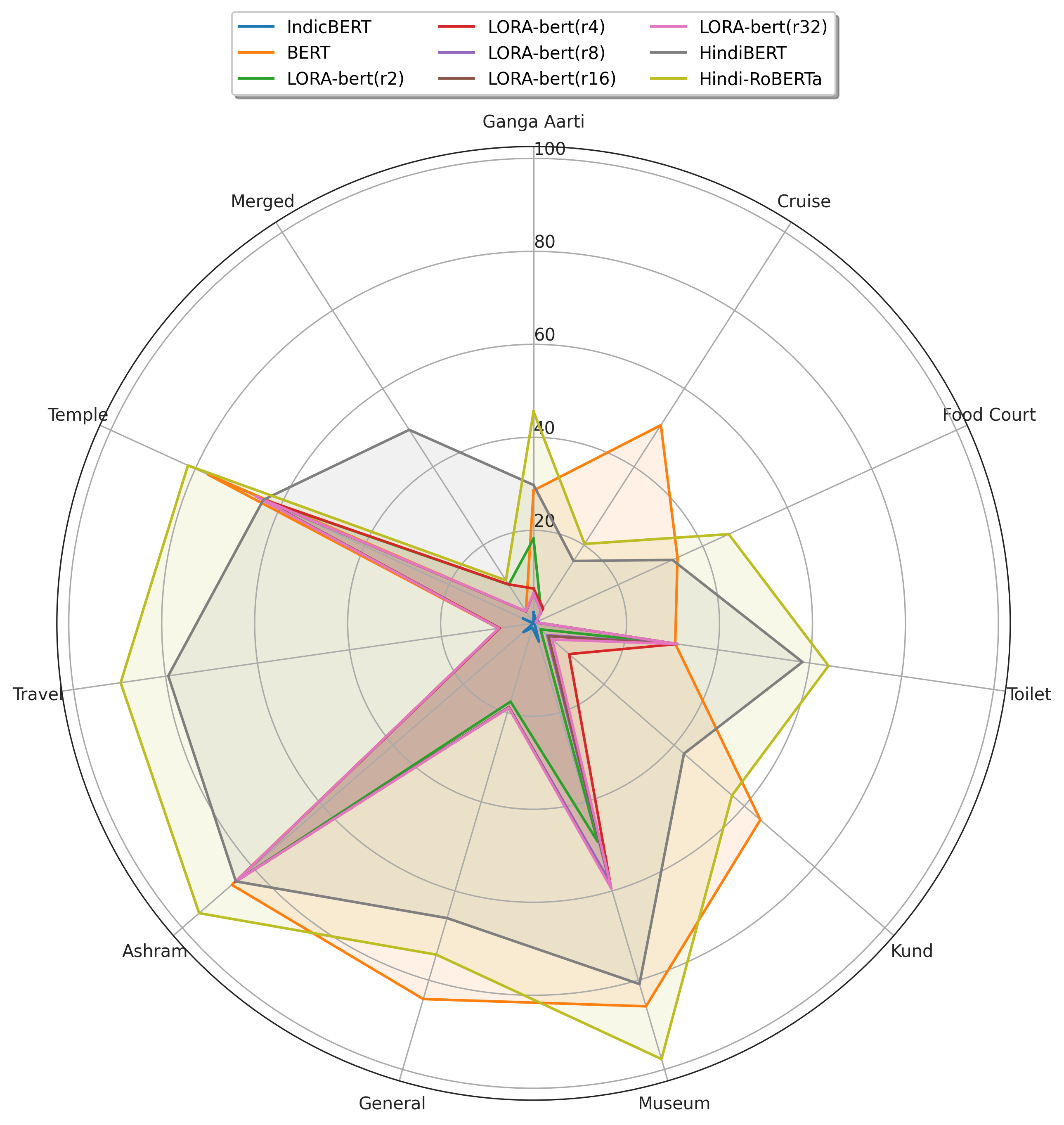}
    \caption{BLEU score comparison of the models across 11 domain settings} 
    \label{fig:bleu_result_compare}
\end{figure}

\begin{figure}[!hbt]
    \centering
    \includegraphics[width=0.5\linewidth]{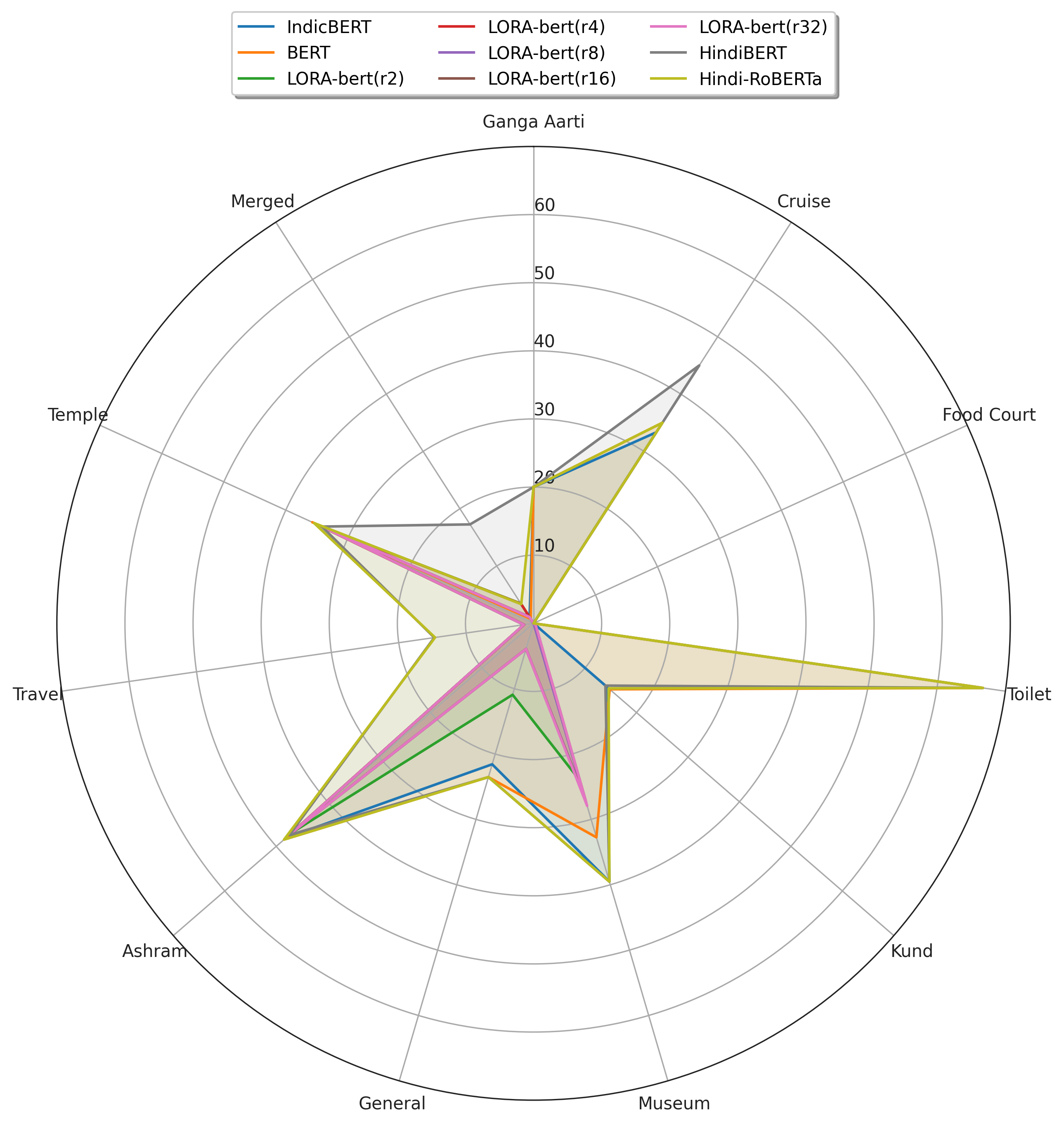}
    \caption{ROUGE-L score comparison of the models across 11 domain settings}
    \label{fig:rouge_result_compare}
\end{figure}

Figure~\ref{fig:result_compare} and \ref{fig:bleu_result_compare} show that models pretrained specifically on Hindi, such as HindiBERT and Hindi-RoBERTa, consistently outperform general-purpose models like BERT and IndicBERT across most domain-specific classification tasks, including Museum, Ashram, Temple, and Travel. Figure~\ref{fig:rouge_result_compare} shows Hindi-RoBERTa outperformed on the Toilet, Museum, Ashram, Temple, and Cruise domains, considering ROUGE-L score.
This underscores the importance of language-specific pretraining for achieving robust performance in Indic NLP. While LoRA-based adaptations (r2 and r4) show improvements over vanilla BERT in selected domains, they still lag behind Hindi-specialized models, indicating that parameter-efficient tuning alone may not suffice without domain-relevant representation learning. IndicBERT, despite its multilingual training, performs poorly overall, suggesting that broader language coverage can dilute effectiveness in Hindi-specific tasks. Notably, the Merged category challenges all general models, with HindiBERT maintaining its advantage, highlighting its robustness in heterogeneous or cross-domain settings.

\section{Conclusion}
Developing extractive QA systems for low-resource languages is particularly challenging in domain-specific settings. In this study, we manually prepared a Hindi QA dataset focused on Varanasi tourism$—$a cultural and spiritual hub renowned for its Bhakti-Bhaav$—$comprising 7,715 question-answer pairs obtained through extensive fieldwork using purposive and convenience sampling, and covering ten tourism-centric subdomains: Ganga Aarti, Cruise, Food Court, Public Toilet, Kund, Museum, General, Ashram, Travel, and Temple. Which was subsequently augmented to 27,455 pairs using a Llama-based zero-shot prompting technique. We propose a framework leveraging foundation models-BERT and RoBERTa, fine-tuned using SFT and LORA, to optimize parameter efficiency and task performance. Multiple variants of BERT, including mBERT, Hindi-BERT, IndicBERT, etc., are evaluated to assess their suitability for low-resource domain-specific QA. In the Ganga Aarti, General, and Toilet sub-domains, the BERT model achieved the highest F1 scores, registering 51.223, 76.216, and 73.469, respectively, while in the Cruise, Kund, Museum, Ashram, Travel, Temple, and Food Court domains, the Hindi-RoBERTa model obtained F1 scores of 46.327, 66.639, 95.532, 96.575, 90.91, 86.493, and 66.987 respectively; as the RoBERTa model was pretrained on a large dataset, fine-tuning on the merged sub-domains enabled the Hindi-RoBERTa model to achieve an overall best F1 score of 89.75. As future work, we will leverage the existing multilingual LLM with RAG to enhance model robustness and effectively handle real-time diversified queries. Additionally, a similar Hindi QA dataset can be developed for other domains, such as education, agriculture, and health sciences.

\appendix

\section{\break Appendix-Sources}
\label{app:1}

\begin{table}[!h]
\centering
\caption{List of sources and links}
\setlength{\tabcolsep}{3pt}
\begin{tabular}{|p{35pt}|p{450pt}|}
\hline
\textbf{Source} & \textbf{Link} \\ \hline

\multirow{7}{*}{Tourism} & \href{http://pawanpath.up.gov.in/}{http://pawanpath.up.gov.in/} \\
 & \href{https://yappe.in/uttar-pradesh/varanasi}{https://yappe.in/uttar-pradesh/varanasi} \\
 & \href{https://uptourism.gov.in/en/page/varanasi-sarnath}{https://uptourism.gov.in/en/page/varanasi-sarnath} \\
 & \href{https://uptourism.gov.in/en/article/year-wise-tourist-statistics}{https://uptourism.gov.in/en/article/year-wise-tourist-statistics} \\
 & \href{https://upstdc.co.in/Web/varanasi_tourism}{https://upstdc.co.in/Web/varanasi\_tourism} \\
 & \href{https://varanasitemples.in/}{https://varanasitemples.in/} \\
 & \href{https://kashiarchan.com/}{https://kashiarchan.com/} \\ \hline
 
\multirow{4}{*}{Articles} & \href{https://www.lonelyplanet.com/articles/guide-to-varanasi}{https://www.lonelyplanet.com/articles/guide-to-varanasi} \\
 & \href{https://travel.india.com/guide/destination/experience-the-magic-of-varanasis-silk-sarees-and-wooden-carvings-7358826/}{https://travel.india.com/guide/destination/experience-the-magic-of-varanasis-silk-sarees-and-wooden-carvings-7358826/} \\
 & \href{https://timesofindia.indiatimes.com/travel/destinations/5-must-visit-places-in-varanasi/articleshow/115978318.cms}{https://timesofindia.indiatimes.com/travel/destinations/5-must-visit-places-in-varanasi/articleshow/115978318.cms} \\
 & \href{https://www.varanasi.org.in/sankatha-temple-varanasi\#google_vignette}{https://www.varanasi.org.in/sankatha-temple-varanasi\#google\_vignette} \\ \hline
 
\multirow{2}{*}{Blogs} & \href{https://classynomad.com/best-street-foods-in-varanasi/}{https://classynomad.com/best-street-foods-in-varanasi/} \\ 
 & \href{https://livingnomads.com/2023/03/varanasi-travel-blog/}{https://livingnomads.com/2023/03/varanasi-travel-blog/} \\ \hline
 
\multirow{2}{*}{Reviews} & \href{https://www.tripadvisor.in/Attractions-g297685-Activities-Varanasi_Varanasi_District_Uttar_Pradesh.html}{https://www.tripadvisor.in/Attractions-g297685-Activities-Varanasi\_Varanasi\_District\_Uttar\_Pradesh.html} \\
 & \href{https://www.tripadvisor.com/Tourism-g297685-Varanasi_Varanasi_District_Uttar_Pradesh-Vacations.html}{https://www.tripadvisor.com/Tourism-g297685-Varanasi\_Varanasi\_District\_Uttar\_Pradesh-Vacations.html} \\ \hline
 
Books & \href{https://www.scribd.com/document/477332597/Shiva-Lingams-of-Kashi}{Shiv Lings of Kashi} \\ \hline

\multirow{3}{*}{YouTube} & \href{https://youtu.be/890MOuBOUGE}{https://youtu.be/890MOuBOUGE} \\
 & \href{https://youtu.be/LLOEVk2FttU?si=PjTJBmYkZjv5eJEY}{https://youtu.be/LLOEVk2FttU?si=PjTJBmYkZjv5eJEY} \\
 & \href{https://youtube.com/shorts/kZM9yQeP-UE?si=gaxSV6Cq16eYBpX4}{https://youtube.com/shorts/kZM9yQeP-UE?si=gaxSV6Cq16eYBpX4} \\ \hline
 
Others & census data, public sector records \\ \hline
\end{tabular}

\label{tab:appendix1}
\end{table}

\section{\break Appendix-Example of semantically repeated QA pairs}
The text in \textcolor{red}{red color} (3-4) represents semantically related QA pairs that were dropped by annotators.
\begin{enumerate}
    \item kyā durgā maṃdira mēṃ tīna bāra āratī hōtī hay? [\footnote{The[] comprising translation of Hindi text}\textit{Does three times aarti take place in Durga Mandir?}]\\
Ans. hāz, durgā maṃdira mēṃ tīna bāra āratī hōtī hay| [\textit{Yes, three times aarti take place in Durga Mandir.}]\\

\item durgā maṃdira mēṃ kitanī bāra āratī hōtī hay? [\textit{How many times does aarti take place at Durga Mandir?}]\\
Ans. durgā maṃdira mēṃ tīna bāra āratī hōtī hay| [\textit{Three times aarti take place at Durga Mandir.}]\\

\item \textcolor{red}{
durgā maṃdira mēṃ kitanī bāra āratī āyōjita hōtī hay? [\textit{How many times is aarti organised at Durga Mandir?}]\\
Ans. durgā maṃdira mēṃ tīna bāra āratī āyōjita hōtī hay| [\textit{Three times aarti is organised at Durga Mandir.}]\\ }

\item \textcolor{red}{durgā maṃdira mēṃ kitanī bāra āratī kā āyōjana hōtā hay? [\textit{How many times is aarti organised at Durga Mandir?}]\\
Ans. durgā maṃdira mēṃ tīna bāra āratī kā āyōjana hōtā hay| [\textit{Three times aarti is organised at Durga Mandir.}]\\ }
\end{enumerate}



\section*{Acknowledgment}
We are grateful to acknowledge Transdisciplinary Research (TDR) Grant as a part of the Institute of Eminence, Banaras Hindu University (BHU), for providing the research grant that made this work in the Varanasi tourism domain possible. We have used DeepSeek and ChatGPT LLMs to refine our methodology and enhance the fluency of the text in the article. We sincerely thank Dr. S. Suresh, NIT, Kurukshetra and Dr. Jagdeesan T., BHU, Varanasi who were Co-Principal Investigators of the TDR Project. Because of their cooperation and coordination during the project tenure, we could work on this task.  We thank Research Assistants of the TDR Project Shreya Pandey, Bhaskar Singh, and Aman Gupta for assisting in the creation of the Hindi QA dataset.  We thank Himesh, and Abhilasha Master's students, BHU for extending their support in compiling the Hindi QA dataset for Varanasi tourism. We also thank Iram Ali Ahmad, Supriya Chauhan, and Jyoti Kumari for helping in refining the Hindi QA dataset.

\bibliographystyle{unsrtnat}
\bibliography{references}  

\end{document}